%% file: acl_latex.tex
\documentclass[11pt]{article}

\usepackage{acl}

\usepackage{times}
\usepackage{latexsym}
\usepackage[T1]{fontenc}
\usepackage[utf8]{inputenc}
\usepackage{microtype}
\usepackage{inconsolata}
\usepackage{graphicx}
\usepackage{amsmath}
\usepackage{amssymb}
\usepackage{booktabs}
\usepackage{stfloats}  

\newcounter{vrow}
\newcommand{\rowlabel}[1]{\refstepcounter{vrow}\thevrow\label{#1}}

\newcommand{\gain}[1]{\textsubscript{$\uparrow$#1}}
\newcommand{\drop}[1]{\textsubscript{$\downarrow$#1}}

\title{Do Value Vectors in Deep Layers Need Context from the Residual Stream?}

\newcommand{\drexel}{\raisebox{-0.15ex}{\includegraphics[height=0.8em]{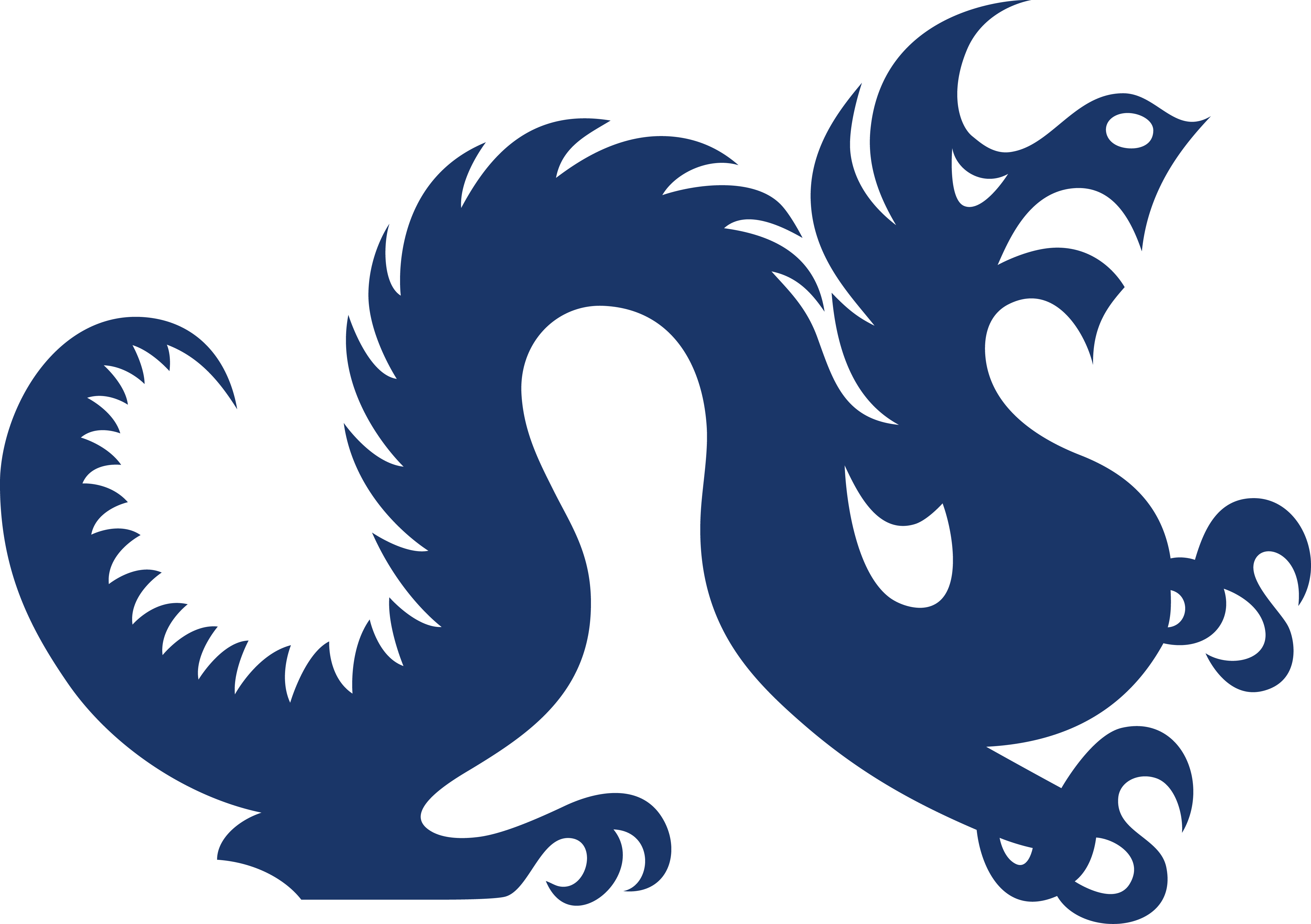}}}
\author{%
  Muyu He\hspace{0.4em}\drexel\textsuperscript{*} \quad
  Yuchen Liu\hspace{0.4em}\drexel\textsuperscript{*} \quad
  Qingya Huang\textsuperscript{\textdagger} \quad
  Li Zhang\hspace{0.4em}\drexel \\[3pt]
  {\mdseries\normalsize
    \drexel\ Drexel University \qquad
    \textsuperscript{\textdagger}Independent \qquad
    \textsuperscript{*}Equal contribution.}%
}

\begin{document}
\maketitle

\begin{abstract}
The success of modern LLMs is in large part due to their use of attention layers.
An attention layer follows the neural network paradigm:
it takes the residual stream as input
and thereby produces context-dependent query, key, and value vectors.
However, we find that model performance meaningfully improves
when deeper layers learn only a context-free value vector to preserve the original token information,
without drawing on any context from the residual stream. 
When the model has access to this context-free value vector, 
adding back the context-dependent component provides little additional benefit for aggregate benchmark performance.
Such context-free value vectors can be stored as sparse model parameters,
eliminating the need to recompute or persistently cache these values.
Through systematic ablations on the key design choices for such context-free value vectors,
we propose \textbf{Bank of Values} (BoV),
a new way of computing value vectors in attention
by learning a lookup table of token-specific value vectors for each of the last third of layers.
Across 135M and 780M models,
BoV improves validation loss over standard attention and,
at 780M, the average score across 21 benchmarks,
matching the previous best method that adds token information to the value vector 
with less compute and memory.\footnote{Full architecture and training code is available at \url{https://github.com/RiddleHe/nanochat}.}
\end{abstract}

\input{sections/introduction}
\input{sections/motivation}
\input{sections/bank_of_values}

\input{sections/experiments}
\input{sections/related_work}
\input{sections/conclusion}
\input{sections/limitations}

\section*{Acknowledgments}
We thank Xiangming Gu for valuable feedbacks on the experiments and presentations of the paper.

\bibliography{custom}

\newpage
\appendix
\input{appendices/additional_results}
\input{appendices/experimental_details}

\end{document}

%% file: sections/introduction.tex
\section{Introduction}

\begin{figure*}[t]
    \centering
    \includegraphics[width=\textwidth]{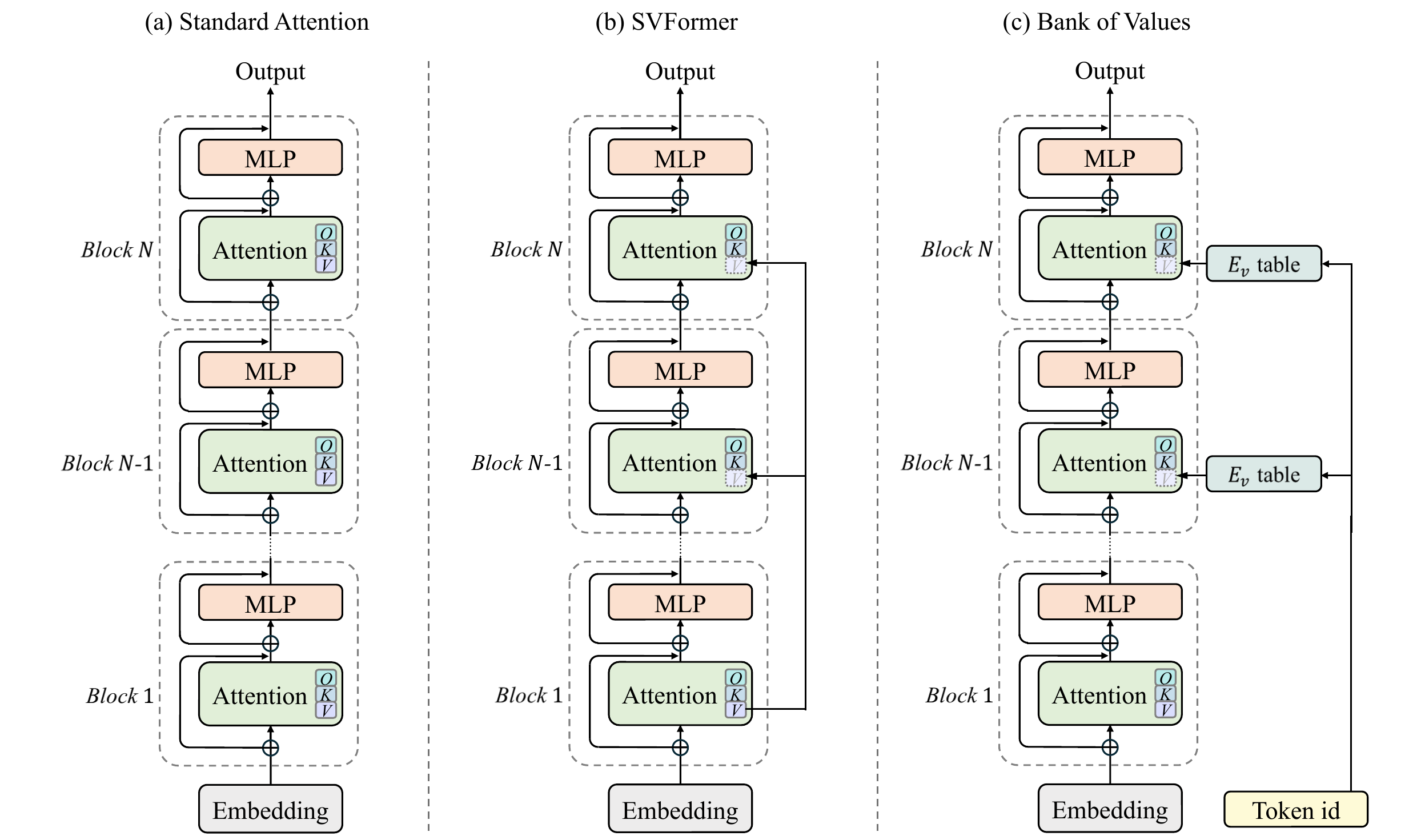}
    \caption{Overview of Bank of Values. (a) Standard Attention: a context-dependent value vector is computed directly from the residual stream. (b) SVFormer: a context-free value vector is copied from the first layer's value vector into all subsequent layers. (c) Bank of Values: in each of the last third of layers, a context-free value vector is looked up by token id from that layer's value vector table $\mathbf{E}_v$ and scaled by a learnable coefficient.}
    \label{fig:arch}
\end{figure*}


Modern LLMs invariably adopt the transformer architecture,
whose basic formulation necessarily includes the attention layer \citep{vaswani2017attention}.
Attention computes dot products between query and key vectors to obtain attention scores,
then writes a weighted sum of value vectors back to the residual stream according to these scores.
Despite its sophisticated design, the transformer retains an old paradigm first proposed in ResNet \citep{he2016resnet}:
the input to each layer of a deep network is the normalized sum of all previous layers' outputs,
which we refer to as the residual stream.
This design proves effective for computing queries and keys,
as it allows deeper layers to accumulate context information
propagated from other tokens through attention \citep{ghandeharioun2024patchscopes}.
However, whether context information is necessary for value vectors is left largely unexamined,
even though value vectors do not themselves participate in cross-token computation.
As a result, the residual stream remains the default input for computing value vectors,
diluting the original context-free token information available to them.


Recent work has begun to explore whether the value vector 
should carry more of the original token information and less of the accumulated context.
Value Residual Learning \citep{zhou2024value} 
linearly interpolates the value vectors in deeper layers with the value vector from the first attention layer.
Since the value vector in the first layer is a linear transformation of the token embedding,
it deposits original token information into the value vectors in deeper layers.
Similar work
modulates value vectors in deeper layers
with a token-specific value vector
that is either shared across layers \citep{li2026move}
or learned separately per layer \citep{karpathy2025nanochat}.
All of these methods claim improvements over the standard attention mechanism,
implying that a larger share of original token information in the value vector benefits deeper layers.
At the same time, studies have explored using only a context-free value vector for attention,
directly reusing the first layer's value vector (SVFormer, \citealp{zhou2024value}), or components of it (SkipV1Former, \citealp{wu2025skipv1former}),
in all subsequent layers.
These methods report degraded performance relative to the standard attention baseline. 
Together, the results suggest that context information from the residual stream is essential
to value vectors in deeper layers,
and that adding a context-free component that preserves the original token information is beneficial but not sufficient.


We find that all the aforementioned studies share the same oversight:
when targeting value vectors in deeper layers,
the effect of computing a single context-free value vector 
is never studied in isolation.
In the rare cases where it is, 
the context-free value vector is nevertheless applied to all layers,
which fails to account for different layers' varying dependence on context.
Systematically ablating how value vectors are computed along four axes
(Section~\ref{subsec:analysis}),
we find that almost all of the reported gains in previous work come from the context-free component
that represents the original token,
not from the context-dependent component.
In these deeper layers, the value can be computed 
without the residual stream at little to no cost to performance.

The insight that deeper layers largely benefit from a context-free value vector
has an important consequence:
since these value vectors do not require context information,
they need not be computed from a particular input sequence,
nor cached for reuse in subsequent decode steps.
Instead, context-free value vectors can persist as FLOP-free, sparse model parameters
that are looked up in $O(1)$ time during a forward pass.
We therefore propose \textbf{Bank of Values} (BoV),
which replaces the standard value vector computation from the residual stream
with a static lookup in a dedicated value vector table in each of the last third of layers.
As a result, these layers drop the value matrix from their parameters and no longer cache value vectors, 
storing only the token indices needed to look the values up.
Although value lookup tables have been explored in the works above, 
they are always added on top of the value computed from the residual stream. 
Consequently, those models still compute and cache that value,
achieving lower quality than BoV at higher memory and compute cost (Section~\ref{subsec:comparison}).


Empirically, we train a transformer model with BoV at two model sizes, 135M and 780M,
on three FLOP budgets, $1.50 \times 10^{18}$, $3.91 \times 10^{19}$, and $4.78 \times 10^{20}$ FLOPs.
In a FLOP-controlled setting, BoV lowers validation loss at both scales 
on a held-out split of roughly 41.9M tokens
from the ClimbMix dataset \citep{diao2025climb} 
and, at 780M, 
raises the average score across the 21 DCLM CORE benchmarks \citep{li2024dclm}, 
relative to an otherwise identical model with standard attention.
Moreover, compared with existing methods that add original token information to the value vector, 
BoV matches the top-performing variant and surpasses the rest, all while using less memory and fewer FLOPs per token.


%% file: sections/motivation.tex
\section{Preliminaries}
\label{sec:preliminaries}


We establish notation for a single forward pass of a transformer model, omitting the batch dimension for clarity.
A model embeds a length-$T$ token sequence into $\mathbf{e} \in \mathbb{R}^{T \times d}$, where $d$ is the hidden dimension.
Its normalized form $\mathbf{x}_0 = \mathrm{RMSNorm}(\mathbf{e})$, which we call the initial token embedding, 
is the input to the first transformer block and carries no context information on its own.

Transformer blocks are joined by residual connections \citep{he2016resnet}.
The input $\mathbf{x}_i$ to the $i$-th block's attention layer, which we call the residual stream,
is the RMSNorm of the sum of the initial token embedding $\mathbf{x}_0$ and the outputs of all preceding layers.
Unlike the context-free $\mathbf{x}_0$, the residual stream $\mathbf{x}_i$ accumulates information from other tokens through the attention layers below it, and is therefore context-dependent.

The attention layer projects query, key, and value vectors from $\mathbf{x}_i$,
\begin{equation}
    \mathbf{Q} = \mathbf{x}_i \mathbf{W}_Q, \quad
    \mathbf{K} = \mathbf{x}_i \mathbf{W}_K, \quad
    \mathbf{V} = \mathbf{x}_i \mathbf{W}_V,
    \label{eq:qkv}
\end{equation}
where $\mathbf{W}_Q, \mathbf{W}_K, \mathbf{W}_V \in \mathbb{R}^{d \times d}$.

Each of $\mathbf{Q}, \mathbf{K}, \mathbf{V}$ is split into $H$ heads of dimension $d_h = d / H$,
and each head $h$ computes scaled dot-product attention,
\begin{equation}
    \mathbf{O}^{(h)} = \mathrm{softmax}\!\left( \frac{\mathbf{Q}^{(h)} {\mathbf{K}^{(h)}}^{\!\top}}{\sqrt{d_h}} \right) \mathbf{V}^{(h)}.
\end{equation}
The per-head outputs are concatenated and projected by $\mathbf{W}_O \in \mathbb{R}^{d \times d}$ into the attention output
$\mathbf{a}_i = \mathrm{concat}_{h=1}^{H}(\mathbf{O}^{(h)})\, \mathbf{W}_O$,
which is added back to the residual stream, updating each token's representation with information from other tokens.

\begin{figure*}[t]
    \centering
    \includegraphics[width=\textwidth]{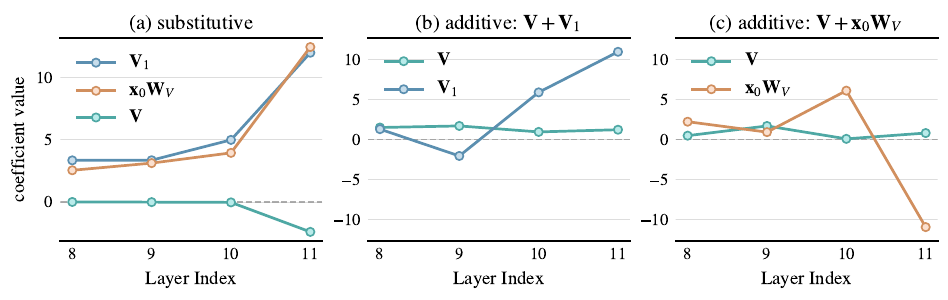}
    \caption{Value of the unbounded, learnable coefficient for each component of the value vector at each target layer of the 12-layer model, for the substitutive (a) and additive (b, c) variants.}
    \label{fig:coeff}
\end{figure*}

\section{Motivation}
\label{sec:motivation}
\label{subsec:analysis}

\input{tables/value_ablation}

In this work,
we comprehensively study to what extent value vectors in attention layers
need context information from the residual stream,
and to what extent they instead benefit from the original token information.
We provide the model with two context-free sources of original token information, which we collectively denote $\tilde{\mathbf{V}}$:
a value vector $\tilde{\mathbf{V}} = \mathbf{x}_0 \mathbf{W}_V$ computed from the initial token embedding $\mathbf{x}_0$
with the target layer's value matrix $\mathbf{W}_V$,
or the value vector $\tilde{\mathbf{V}} = \mathbf{V}_1$ from the first attention layer.
In each variant, $\tilde{\mathbf{V}}$ either replaces or is linearly combined with
the standard value vector $\mathbf{V} = \mathbf{x}_i \mathbf{W}_V$ computed from the residual stream.

Our exploration spans four axes.
(1) \textbf{Additive or substitutive.}
We study whether the model benefits more from using $\tilde{\mathbf{V}}$ directly,
which substitutes $\mathbf{V}$ entirely and supplies only context-free, original token information,
or from a weighted combination of $\tilde{\mathbf{V}}$ and $\mathbf{V}$,
which mixes context-free and context-dependent information.
Unless otherwise noted, all variants modify only the last third of attention layers.
(2) \textbf{Learnable or fixed coefficient.}
We study whether each component of the value vector
is best scaled by an unbounded, learnable coefficient or by a fixed one.
In the substitutive variants, the fixed coefficient for $\tilde{\mathbf{V}}$ is $1$;
in the additive variants, the fixed coefficient for both $\tilde{\mathbf{V}}$ and $\mathbf{V}$ is $0.5$.
The magnitudes of the learned coefficients reveal how strongly the model prefers $\tilde{\mathbf{V}}$ over $\mathbf{V}$.
(3) \textbf{Source of original token information.}
We study using $\mathbf{x}_0 \mathbf{W}_V$, which is unique to each target layer,
or $\mathbf{V}_1$, which is shared across all target layers, as $\tilde{\mathbf{V}}$.
(4) \textbf{Target layers.}
We study the effect of targeting the last third of layers, every other layer, or every layer.

We conduct all experiments on a 12-layer, 135M transformer model
in the style of GPT-2 \citep{radford2019language},
using a simplified implementation of nanochat \citep{karpathy2025nanochat}.
We train every variant under a compute-controlled budget of $1.50 \times 10^{18}$ FLOPs,
which yields 1.97B tokens for the standard attention baseline.\footnote{This budget is cost-effective: its validation loss, measured in bits per byte (BPB), is only 0.06 BPB higher than that of training on 40B tokens.}
All runs use a fixed global batch size of 524{,}288 tokens and a sequence length of 2048.
We use the Muon optimizer for all matrix parameters with a learning rate of $0.02$,
and the AdamW optimizer for the remaining parameters,
following nanochat's standard practices.
A complete training configuration is given in Appendix~\ref{app:details}.

From Table~\ref{tab:value_ablation} and Figure~\ref{fig:coeff}, we make the following observations
about the relative contributions of $\mathbf{V}$ and $\tilde{\mathbf{V}}$ to the value vector.

(i) \textbf{Deep layers prefer a larger share of $\tilde{\mathbf{V}}$ over $\mathbf{V}$.}
Figure~\ref{fig:coeff}(b, c) shows that, in the two additive variants,
the learned coefficients for $\tilde{\mathbf{V}}$ are much larger in absolute value than those for $\mathbf{V}$.
Several layers drive the coefficient of $\mathbf{V}$ toward $0$, giving it minimal influence,
while the final layer pushes the magnitude of $\tilde{\mathbf{V}}$'s coefficient past $10$.
When the value vector is computed from a single source, Figure~\ref{fig:coeff}(a) shows that
the coefficients for $\tilde{\mathbf{V}}$ stay consistently above $2$ and rise past $10$ in the final layer,
while the coefficient for $\mathbf{V}$ is driven toward $0$ or below.
Fixing the coefficients for both components in the additive variants
at $0.5$ (rows~\ref{row:v1-add-half} and~\ref{row:x0-add-half})
increases the validation loss,
suggesting that the model benefits from an uneven weighting
in favor of $\tilde{\mathbf{V}}$.

(ii) \textbf{Only deep layers, not shallow ones, benefit from substituting $\mathbf{V}$ with $\tilde{\mathbf{V}}$.}
With a fixed coefficient of $1$, substituting $\mathbf{V}$ with $\mathbf{V}_1$ in every layer (row~\ref{row:v1-all})
leads to a higher loss than the standard attention baseline (row~\ref{row:base}),
whereas restricting the substitution to the last third of layers achieves significantly lower loss (row~\ref{row:v1-last}).
With a learnable coefficient, substituting $\mathbf{V}$ with $\mathbf{V}_1$ (row~\ref{row:v1-every2})
or $\mathbf{x}_0\mathbf{W}_V$ (row~\ref{row:x0-every2}) in every other layer
still lags behind substituting in the last third of layers (rows~\ref{row:v1-last-c} and~\ref{row:x0-last-c}).
Substituting $\mathbf{V}$ with $\tilde{\mathbf{V}}$
therefore provides little benefit in the shallow layers
but a meaningful improvement in the deep ones.

(iii) \textbf{Once deep layers have access to $\tilde{\mathbf{V}}$, adding $\mathbf{V}$ yields little improvement.}
With learnable coefficients, computing the value vector directly as $\mathbf{V}_1$ (row~\ref{row:v1-last-c})
achieves a slightly lower loss than $\mathbf{V} + \mathbf{V}_1$ (row~\ref{row:v1-add}),
while computing it directly as $\mathbf{x}_0\mathbf{W}_V$ (row~\ref{row:x0-last-c})
matches the loss of $\mathbf{V} + \mathbf{x}_0\mathbf{W}_V$ (row~\ref{row:x0-add}).
This corroborates observation (i): the contribution of $\mathbf{V}$ is minimized at deep layers in the additive variants.

(iv) \textbf{Deep layers benefit from a layer-specific value vector rather than a shared one.}
With both a fixed and a learnable coefficient, using a layer-specific $\mathbf{x}_0\mathbf{W}_V$ as $\tilde{\mathbf{V}}$
(rows~\ref{row:x0-last} and~\ref{row:x0-last-c})
consistently outperforms using a shared $\mathbf{V}_1$ (rows~\ref{row:v1-last} and~\ref{row:v1-last-c}).
This is consistent with $\tilde{\mathbf{V}} = \mathbf{x}_0\mathbf{W}_V$ being strictly more expressive than $\tilde{\mathbf{V}} = \mathbf{V}_1$:
although both are computed from the same input $\mathbf{x}_0$,
multi-head attention lets each layer's value matrix $\mathbf{W}_V$ project $\mathbf{x}_0$ into a different low-rank subspace,
which a single shared projection cannot reproduce.

Overall, when restricted to the last third of layers, computing the value vector directly from a layer-specific $\mathbf{x}_0\mathbf{W}_V$ provides the largest gain,
and the role of $\mathbf{V}$ is marginal.
This shows that value vectors in deep layers benefit primarily from context-free, original token information,
an insight we build on in our architecture design in Section~\ref{sec:bov}.

%% file: tables/value_ablation.tex
\begin{table}[t]
    \centering
    \small
    \setcounter{vrow}{0}
    \begin{tabular*}{\columnwidth}{@{\extracolsep{\fill}}clccc@{}}
        \toprule
        & Output & Target & Coeff. & Val.\ loss \\
        \midrule
        \rowlabel{row:base}     & $\mathbf{V}$ & ---    & ---          & 0.854 \\
        \rowlabel{row:base-c}   & $\mathbf{V}$ & Last 4 & $\gamma_v$   & 0.858 \\
        \midrule
        \rowlabel{row:v1-all}     & $\mathbf{V}_1$ & All 12    & $1$        & 0.872 \\
        \rowlabel{row:v1-every2}  & $\mathbf{V}_1$ & Every 2   & $\gamma_v$ & 0.851 \\
        \rowlabel{row:v1-last}    & $\mathbf{V}_1$ & Last 4    & $1$        & 0.849 \\
        \rowlabel{row:v1-last-c}  & $\mathbf{V}_1$ & Last 4    & $\gamma_v$ & 0.847 \\
        \midrule
        \rowlabel{row:x0-every2}  & $\mathbf{x}_0\mathbf{W}_V$ & Every 2   & $\gamma_v$ & 0.849 \\
        \rowlabel{row:x0-last}    & $\mathbf{x}_0\mathbf{W}_V$ & Last 4    & $1$        & 0.847 \\
        \rowlabel{row:x0-last-c}  & $\mathbf{x}_0\mathbf{W}_V$ & Last 4    & $\gamma_v$ & \textbf{0.845} \\
        \midrule
        \rowlabel{row:v1-add-half} & $\mathbf{V} + \mathbf{V}_1$ & Last 4 & $0.5$      & 0.849 \\
        \rowlabel{row:v1-add}      & $\mathbf{V} + \mathbf{V}_1$ & Last 4 & $\gamma_v$ & 0.848 \\
        \rowlabel{row:x0-add-half} & $\mathbf{V} + \mathbf{x}_0\mathbf{W}_V$ & Last 4 & $0.5$      & 0.848 \\
        \rowlabel{row:x0-add}      & $\mathbf{V} + \mathbf{x}_0\mathbf{W}_V$ & Last 4 & $\gamma_v$ & \textbf{0.845} \\
        \bottomrule
    \end{tabular*}
    \caption{Validation loss (BPB) on the 12-layer, 135M model across variants. $\gamma_v$ indicates an unbounded, learnable coefficient.}
    \label{tab:value_ablation}
\end{table}

%% file: sections/bank_of_values.tex
\section{Bank of Values: Persisting Context-Free Value Vectors as Model Parameters}
\label{sec:bov}


\begin{figure*}[t]
    \centering
    \includegraphics[width=\textwidth]{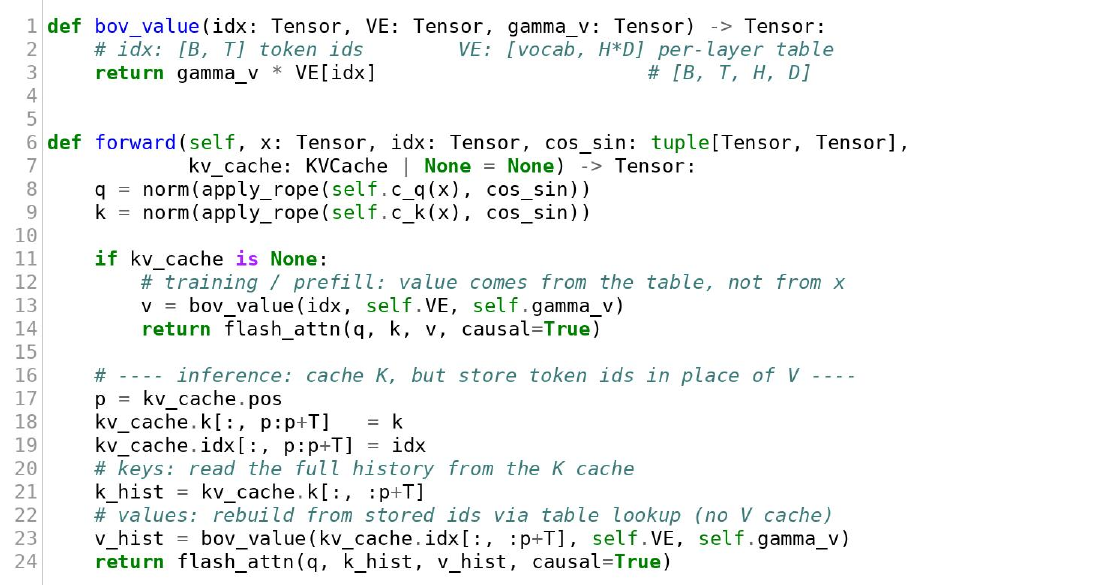}
    \caption{PyTorch-style pseudocode for Bank of Values. \texttt{bov\_value()} performs an $O(Td)$ lookup of all past value vectors, scaled by a coefficient \texttt{gamma\_v}. \texttt{forward()} stores past token indices in \texttt{kv\_cache.idx} and uses them to copy value vectors into \texttt{v\_hist} for dot-product attention.}
    \label{fig:pseudocode}
\end{figure*}

From Section~\ref{sec:motivation}, value vectors in deep attention layers
are most effective when computed directly from $\mathbf{x}_0\mathbf{W}_V$.
Importantly, since both $\mathbf{x}_0$ and $\mathbf{W}_V$ depend only on model parameters,
the value vector that a token with id $i$ produces,
\begin{equation}
    \mathbf{v}_i = \mathrm{RMSNorm}(\mathbf{e}_i)\,\mathbf{W}_V,
    \label{eq:detval}
\end{equation}
is deterministic and independent of the preceding sequence, where $\mathbf{e}_i$ is the embedding of the token with id $i$.
We can therefore learn each $\mathbf{v}_i$ directly as a model parameter, stored at each target layer in a value table
\begin{equation}
    \mathbf{E}_v \in \mathbb{R}^{|\mathcal{V}| \times d}, \qquad \mathbf{E}_v[i] = \mathbf{v}_i,
    \label{eq:evtable}
\end{equation}
where $|\mathcal{V}|$ is the vocabulary size and $d$ the hidden dimension.
During a forward pass, the value vector of each token in the last third of layers
is looked up directly from $\mathbf{E}_v$.
Scaling each looked-up value by an unbounded, learnable per-layer coefficient $\gamma_v \in \mathbb{R}$,
we obtain an attention variant that replaces the dense value matrix $\mathbf{W}_V$
with a sparse lookup table $\mathbf{E}_v$, computing the value vector at position $p$ as
\begin{equation}
    \mathbf{v}_p = \gamma_v\, \mathbf{E}_v[i_p],
    \label{eq:bov}
\end{equation}
where $i_p$ is the id of the token at position $p$.
We call this attention variant \textbf{Bank of Values} (BoV). 
The architecture of BoV is shown in Figure~\ref{fig:arch},
alongside standard attention and SVFormer \citep{zhou2024value}.

\paragraph{Inference.}
In standard attention, the model caches the value vectors of all previous tokens during decode to avoid recomputing them,
reducing the per-step cost of producing past values from $O(Td^2)$ to $O(d^2)$.
In BoV, by contrast, the value vector of each token is looked up directly from $\mathbf{E}_v$,
so no projection or recomputation is ever required, and the value vectors of all $T$ preceding tokens are retrieved by a single \texttt{gather} in $O(Td)$ time.
When a target layer computes attention, the model gathers the value vectors of all preceding tokens from $\mathbf{E}_v$
and discards them as soon as the layer finishes.
Across the target layers, therefore, at most one layer materializes the full set of value vectors at any moment,
in contrast to standard attention, which must keep a value cache for every layer at all times.

\paragraph{Memory overhead.}
Because BoV persists the value vectors of all vocabulary tokens as the parameter table $\mathbf{E}_v$,
the fixed memory cost of $\mathbf{E}_v$ trades off predictably against the value cache of standard attention across context lengths.
The net V-cache memory saved across all layers is
\begin{equation}
    \Delta_{\mathrm{KV}} = \frac{L}{3}\, d \,(T - |\mathcal{V}|),
    \label{eq:kvsave}
\end{equation}
where $L$ is the number of layers, $T$ the context length, $d$ the hidden dimension, and $|\mathcal{V}|$ the vocabulary size.
As Eq.~\ref{eq:kvsave} shows, standard attention is more memory-efficient at shorter contexts,
but beyond the breakeven length $T = |\mathcal{V}|$, BoV saves memory, with the saved fraction growing and eventually saturating.
Figure~\ref{fig:memsave_ablation}(a) gives an overview of this tradeoff for the vocabulary sizes of two tokenizers (nanochat and Qwen3).

For the workloads of modern LLMs, we argue that BoV is preferable for two reasons.
First, long-context inference is a common requirement across a wide range of LLM applications,
precisely the regime in which persisting value vectors as parameters becomes advantageous.
Second, the lookup into $\mathbf{E}_v$ requires only the token indices $i_p$,
which are available well before the forward pass reaches the target layer.
Since only entries corresponding to those token indices need to be retrieved
in a forward pass,
$\mathbf{E}_v$ is a sparse parameter that does not participate in dense matrix multiplication.
Therefore, $\mathbf{E}_v$ can be offloaded to host memory by default,
and only the needed entries are prefetched on demand, improving memory utilization.

We provide PyTorch-style pseudocode for BoV in Figure~\ref{fig:pseudocode}.

\begin{figure*}[t]
    \centering
    \includegraphics[width=\textwidth]{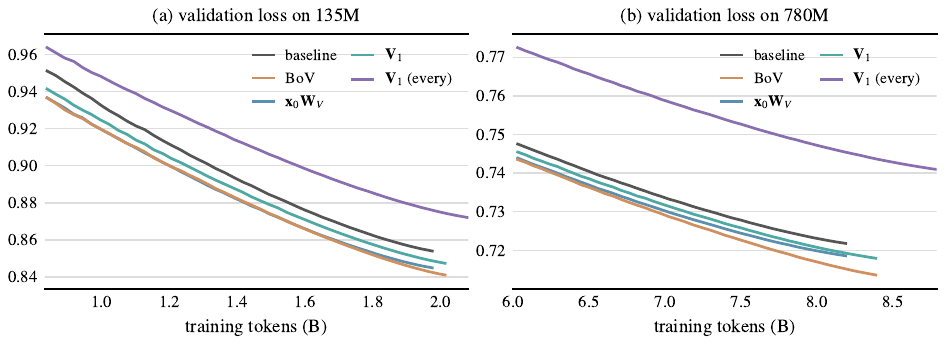}
    \caption{Validation loss over training at the (a) 135M and (b) 780M scales, for the standard attention baseline, BoV, and other attention variants that at least partially compute value vectors from a context-free component.}
    \label{fig:valbpb}
\end{figure*}

%% file: sections/experiments.tex
\section{Experiments}
\label{sec:experiments}

\input{tables/main_100b}
\input{tables/main_results}
\input{tables/add_vs_replace}


\paragraph{Model architecture and training setup.}
We train both BoV and the standard attention baseline at the 24-layer, 780M scale
in the style of nanochat \citep{karpathy2025nanochat}.
We fix the token budget for both at 100B tokens,
which corresponds to $4.78\times10^{20}$ FLOPs for the baseline
and a slightly smaller $4.66\times10^{20}$ for BoV.
As in Section~\ref{sec:motivation},
we use the Muon optimizer with learning rate $0.02$ on all matrix parameters,
and the AdamW optimizer with learning rate $0.15$
on the value vector table $\mathbf{E}_v$ in BoV.
The global batch size is fixed at 1{,}048{,}576 tokens with a sequence length of 2048.
A complete training configuration can be found in Appendix~\ref{app:details}.
We additionally train both architectures at the 12-layer, 135M scale,
following the recipe in Section~\ref{subsec:analysis}.

To avoid introducing any inductive bias through BoV's initialization,
we initialize each target layer's table by copying $\mathbf{x}_0\,\mathbf{W}_V$,
evaluated at each token, into the corresponding row $\mathbf{E}_v[i]$,
so that BoV's first forward pass is numerically identical to an attention layer
that computes its value vectors from the initial token embedding alone.

\paragraph{Evaluation setup.} 
For the 24-layer variants, 
we evaluate on 21 benchmarks from the DCLM CORE suite \citep{li2024dclm}:
Jeopardy~\citep{li2024dclm}, ARC-Easy and ARC-Challenge~\citep{clark2018arc},
COPA~\citep{roemmele2011copa}, CommonsenseQA~\citep{talmor2019commonsenseqa}, PIQA~\citep{bisk2020piqa}, OpenBookQA~\citep{mihaylov2018openbookqa},
LAMBADA~\citep{paperno2016lambada}, HellaSwag~\citep{zellers2019hellaswag}, Winograd~\citep{levesque2012winograd}, Winogrande~\citep{sakaguchi2019winogrande},
AGIEval LSAT-AR~\citep{zhong2023agieval}, SQuAD~\citep{rajpurkar2016squad}, CoQA~\citep{reddy2019coqa}, BoolQ~\citep{clark2019boolq},
and the BIG-bench QA WikiData, Language Identification, Dyck Languages, CS Algorithms, Operators, and Repeat-Copy-Logic tasks~\citep{srivastava2022bigbench}.

Following DCLM, we report a centered accuracy for each task,
$(\text{acc} - r)/(1 - r)$ where $r$ is the task's random-guessing (or majority-class) baseline, 
and define the aggregate CORE score as the unweighted mean of the 21 centered accuracies. 
For both the 12-layer and 24-layer variants, 
we additionally report bits-per-byte validation loss 
on a held-out split of the ClimbMix pretraining corpus, 
computed over approximately 41.9M tokens. 
The main results, 
including scores on four representative benchmarks,  
are reported in Table~\ref{tab:main_results},
and the full benchmark results in Table~\ref{tab:full_results}.

\subsection{Main Results}

As shown in Figure~\ref{fig:valbpb}(a, b) and Table~\ref{tab:main_100b},
BoV outperforms the standard attention baseline
on validation loss at both the 135M and 780M scales.
For the 780M models trained on 100B tokens,
BoV also performs favorably
on 16 of 21 benchmarks (Tables~\ref{tab:main_100b} and~\ref{tab:full_results_24l}),
with the largest gains on commonsense reasoning (CommonsenseQA, COPA)
and symbolic-problem-solving (BIG-bench) tasks,
while showing improvement on most reading comprehension, world knowledge, and language understanding tasks.

\subsection{Comparison with Prior Methods}
\label{subsec:comparison}
We compare BoV with the prior methods discussed in Section~\ref{sec:motivation}
that compute the value vector at least partially from a context-free component $\tilde{\mathbf{V}}$ (Table~\ref{tab:main_results}).
We train all models under an identical compute budget of $3.91\times10^{19}$ FLOPs,
which corresponds to 8.39B tokens for BoV.
At the 780M scale, none of these methods outperforms the standard attention baseline on the aggregate CORE score,
indicating limited scalability.
In particular, computing $\mathbf{x}_0\mathbf{W}_V$ per layer (row~\ref{row:m-x0}),
which has the same expressivity as BoV,
lags behind on benchmark performance.
We surmise that one contributor to this gap is
BoV's lower FLOPs per token, which lets it see more tokens under a FLOP-controlled setting.

In Table~\ref{tab:add_replace}, we also compare BoV with a popular additive variant
\citep{jordan2024moddednanogpt, snimu2025valueemb}
that adds a value-embedding lookup $\mathbf{E}_v[i]$ to the standard value vector computed from $\mathbf{x}_i$.
We follow their original implementation, using input-dependent, bounded scaling coefficients and targeting every second layer.
BoV performs favorably on most benchmarks while attaining a similar validation loss.
The full comparison is reported in Table~\ref{tab:nanogpt_full}.

\begin{figure*}[t]
    \centering
    \includegraphics[width=\textwidth]{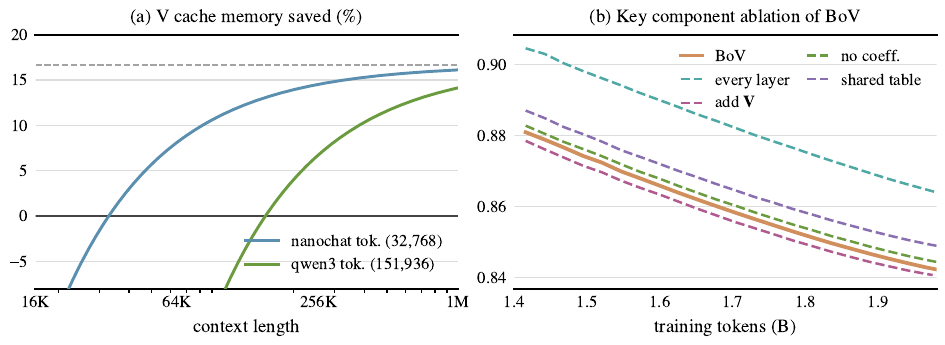}
    \caption{(a) Fraction of V cache memory saved by Bank of Values over standard attention as a function of context length, for the nanochat and Qwen3 tokenizers; and (b) ablation of BoV's design components on the 135M model.}
    \label{fig:memsave_ablation}
\end{figure*}

\subsection{Component Design}

Finally, we ablate each key component of BoV on the 12-layer model
and report the impact on validation loss in Figure~\ref{fig:memsave_ablation}(b).

\paragraph{Sharing a single table across layers.}
Sharing a single $\mathbf{E}_v$ across all target layers increases validation loss,
showing the importance of learning a layer-specific table of value vectors.

\paragraph{Fixing the coefficient at $1$.}
Removing the learnable coefficient and fixing it at $1$ also increases validation loss,
indicating the benefit of scaling the attention output of the target layers.

\paragraph{Targeting every layer.}
Learning an $\mathbf{E}_v$ at every layer rather than only the last third
increases validation loss significantly,
corroborating the insight that only deep layers benefit substantially from context-free value vectors.

\paragraph{Retaining the standard value $\mathbf{V}$.}
Preserving the standard value vector $\mathbf{V}$ computed from the residual stream
yields only a marginal improvement over BoV,
showing that a context-invariant value vector is sufficient in deeper layers.

%% file: tables/main_100b.tex
\begin{table*}[t]
    \centering
    \small
    \setlength{\tabcolsep}{4.5pt}
    \begin{tabular}{l|cc|ccccccc}
        \toprule
        & val bpb~$\downarrow$ & CORE~$\uparrow$ & CSQA & SQuAD & CoQA & Jeopardy & HSwag & WGrande & LAMBADA \\
        \midrule
        Baseline        & 0.671 & 0.334 & 0.437 & 0.506 & 0.342 & 0.219 & 0.660 & 0.581 & 0.504 \\
        \textbf{BoV}    & \textbf{0.663}\drop{.008} & \textbf{0.351}\gain{.017} & \textbf{0.523}\gain{.086} & \textbf{0.531}\gain{.025} & \textbf{0.372}\gain{.030} & \textbf{0.254}\gain{.035} & \textbf{0.669}\gain{.009} & \textbf{0.609}\gain{.028} & \textbf{0.521}\gain{.017} \\
        \bottomrule
    \end{tabular}
    \caption{Validation loss, aggregate CORE score, and representative benchmark scores for standard attention and BoV on the 780M model, trained identically on 100B tokens.}
    \label{tab:main_100b}
\end{table*}

%% file: tables/main_results.tex
\begin{table*}[t]
    \centering
    \small
    \setlength{\tabcolsep}{5pt}
    \setcounter{vrow}{0}
    \begin{tabular}{clcc|cccccc}
        \toprule
        & Output & Layers & Coeff. & val bpb~$\downarrow$ & CORE~$\uparrow$ & ARC-E & HSwag & SQuAD & CoQA \\
        \midrule
        \rowlabel{row:m-base}~(baseline) & $\mathbf{V}$ & --- & --- & 0.722 & 0.260 & 0.691 & 0.558 & 0.396 & 0.291 \\
        \rowlabel{row:m-bov}~(\textbf{BoV}) & $\mathbf{E}_v[i]$ & last 1/3 & $\gamma_v$ & \textbf{0.714}\drop{.008} & \textbf{0.272}\gain{.012} & \textbf{0.716}\gain{.025} & \textbf{0.574}\gain{.016} & \textbf{0.416}\gain{.020} & \textbf{0.321}\gain{.030} \\
        \midrule
        \multicolumn{10}{c}{Input vector variants} \\
        \midrule
        \rowlabel{row:m-x0} & $\mathbf{x}_0\mathbf{W}_V$ & last 1/3 & $\gamma_v$ & 0.719\drop{.003} & 0.259\drop{.001} & 0.714\gain{.023} & 0.562\gain{.004} & 0.396 & 0.297\gain{.006} \\
        \rowlabel{row:m-v1} & $\mathbf{V}_1$ & last 1/3 & $\gamma_v$ & 0.718\drop{.004} & 0.242\drop{.018} & 0.702\gain{.011} & 0.566\gain{.008} & 0.380\drop{.016} & 0.298\gain{.007} \\
        \midrule
        \multicolumn{10}{c}{Target layer variants} \\
        \midrule
        \rowlabel{row:m-v1every} & $\mathbf{V}_1$ & every & $1$ & 0.741\gain{.019} & 0.210\drop{.050} & 0.677\drop{.014} & 0.516\drop{.042} & 0.230\drop{.166} & 0.229\drop{.062} \\
        \bottomrule
    \end{tabular}
    \caption{Validation loss, aggregate CORE score, and representative benchmark scores for standard attention, BoV, and other variants on the 780M model, trained identically on $3.91\times10^{19}$ FLOPs.}
    \label{tab:main_results}
\end{table*}

%% file: tables/add_vs_replace.tex
\begin{table}[t]
    \centering
    \small
    \begin{tabular*}{\columnwidth}{@{\extracolsep{\fill}}lcccc@{}}
        \toprule
        & HSwag & LAMBADA & CSQA & BoolQ \\
        \midrule
        $\mathbf{E}_v[i]$ & \textbf{0.574} & \textbf{0.451} & \textbf{0.324} & \textbf{0.635} \\
        $\mathbf{V} + \mathbf{E}_v[i]$ & \textbf{0.574} & 0.433 & 0.259 & 0.594 \\
        \bottomrule
    \end{tabular*}
    \caption{Comparison on four representative benchmarks between BoV (top), which retrieves value vectors as $\mathbf{E}_v[i]$, and an additive variant (bottom), which adds the retrieved values to the standard value vector as $\mathbf{V} + \mathbf{E}_v[i]$, on the 780M model. The better result in each column is in \textbf{bold}.}
    \label{tab:add_replace}
\end{table}

%% file: sections/related_work.tex
\section{Related Work}

\paragraph{Adding early-layer information to value vectors.}
Several methods inject early-layer information into the value vectors of deeper layers.
Value Residual Learning adds or shares the first layer's values \citep{zhou2024value},
SkipV1Former reuses its value heads \citep{wu2025skipv1former},
and modded-nanogpt \citep{jordan2024moddednanogpt} and MoVE \citep{li2026move} add a per-layer token value-embedding table to the value vector;
earlier sequence models similarly let layers attend to combinations of previous representations \citep{bapna2018training}.
Preserving early-layer information is also motivated by over-smoothing in deep layers \citep{nguyen2023neutreno}.
BoV instead \emph{replaces} the value vector in deep layers with a layer-specific lookup keyed by token identity.

\paragraph{Depth-aware transformer architectures.}
Another line changes how information flows across depth.
DenseFormer feeds each block a learned weighted average of all previous block outputs \citep{pagliardini2024denseformer},
Hyper-Connections replace the fixed residual connection with learnable connection weights \citep{zhu2024hyperconnections},
and Attention Residuals replace the additive residual with a learned, input-dependent weighted sum over previous layer outputs \citep{kimi2026attentionresiduals}.
Mixture-of-Depths Attention further lets each attention head attend to key/value pairs from preceding layers, not just the current one \citep{zhu2026moda}.
BoV instead modifies only the value vectors, leaving the rest of attention unchanged.

\paragraph{Parametric memory and lookup-based computation.}
A related line replaces learned matrix computations with table lookups.
Persistent-memory attention augments each layer with static, learnable key/value vectors \citep{sukhbaatar2019};
product-key memory adds a large key--value lookup layer retrieved via factored keys \citep{lample2019pkm};
MemoryFormer replaces a transformer's linear projection layers with hashed embedding lookups \citep{ding2024memoryformer};
and Engram offloads static information into a sparse, conditional $O(1)$ lookup \citep{cheng2026engram}.
In theory, one of the query, key, or value projections can even be dropped without loss of expressivity \citep{karbevski2025kvweights}.
BoV applies this idea to the value path, replacing the value projection in deep layers with a per-token learned table.

\paragraph{Efficient attention and KV-cache reduction.}
Many methods improve efficiency by optimizing attention kernels \citep{dao2022flashattention, dao2023flashattention2, shah2024flashattention3},
cache layout and serving \citep{kwon2023pagedattention},
sparsifying attended tokens \citep{yuan2025nsa, deepseekai2025deepseekv32},
sharing or compressing KV heads \citep{shazeer2019mqa, ainslie2023gqa, deepseekv2, brandon2024cla},
quantizing the cache \citep{hooper2024kvquant, liu2024kivi},
or pruning cached layers and states \citep{sun2024yoco, wu2024lckv, shen2025lava};
others eliminate the V cache by recomputing values from the residual stream on demand \citep{qasim2026residual}.
These change how attention is computed or which states are kept; BoV instead removes the need to compute or store deep-layer values at all, and is composable with them.

%% file: sections/conclusion.tex
\section{Conclusion}

In this work, we systematically study the value of adding context-free, original token information to value vectors.
We find that deeper layers benefit largely from a context-free value vector that does not draw on the residual stream,
which leads us to propose Bank of Values, an attention variant that persists value vectors directly as model parameters.
Across downstream benchmarks and validation loss, Bank of Values outperforms the standard attention baseline
and scales better than competing variants, while using less memory and fewer FLOPs.

%% file: sections/limitations.tex
\section*{Limitations}

Our study has several limitations.
First, owing to compute constraints, we validate Bank of Values only at the 135M and 780M scales under FLOP-controlled budgets;
confirming that the gains persist for substantially larger models trained on more tokens is left to future work.
Second, our findings are empirical: we observe that only deeper layers benefit from a context-free value vector,
but we do not fully characterize the mechanism behind this depth dependence or the optimization objective it serves,
which we believe is an important direction for understanding the role of value vectors in attention.
Finally, the memory benefit of Bank of Values is context-length dependent: it stores a fixed value table of size $O(|\mathcal{V}|\,d)$ per target layer
and only saves memory beyond a breakeven context length, so its advantage is realized primarily in long-context settings.

%% file: appendices/additional_results.tex
\section{Additional Results}
\label{app:results}

Tables~\ref{tab:full_results}, \ref{tab:full_results_24l}, and~\ref{tab:nanogpt_full} report the full benchmark results behind Tables~\ref{tab:main_results}, \ref{tab:main_100b}, and~\ref{tab:add_replace}, respectively.

\input{tables/full_results}
\input{tables/full_results_24l}
\input{tables/nanogpt_full}

%% file: tables/full_results.tex
\begin{table}[t]
    \centering
    \small
    \begin{tabular*}{\columnwidth}{@{\extracolsep{\fill}}lcc@{}}
        \toprule
        & Baseline & BoV \\
        \midrule
        val bpb~$\downarrow$ & 0.722 & \textbf{0.714}\drop{.008} \\
        CORE~$\uparrow$      & 0.260 & \textbf{0.272}\gain{.012} \\
        \midrule
        \multicolumn{3}{c}{Reading comprehension} \\
        \midrule
        SQuAD                & 0.396 & \textbf{0.416}\gain{.020} \\
        CoQA                 & 0.291 & \textbf{0.321}\gain{.030} \\
        BoolQ                & 0.619 & \textbf{0.635}\gain{.016} \\
        \midrule
        \multicolumn{3}{c}{Symbolic problem solving} \\
        \midrule
        BIG-bench Dyck       & 0.121 & \textbf{0.147}\gain{.026} \\
        AGIEval LSAT-AR      & 0.265 & \textbf{0.291}\gain{.026} \\
        BIG-bench CS-Alg.\   & 0.411 & \textbf{0.433}\gain{.022} \\
        BIG-bench Operators  & 0.181 & \textbf{0.195}\gain{.014} \\
        BIG-bench Rep.-Copy  & 0.000 & \textbf{0.063}\gain{.063} \\
        \midrule
        \multicolumn{3}{c}{World knowledge} \\
        \midrule
        Jeopardy             & 0.082 & \textbf{0.119}\gain{.037} \\
        BIG-bench WikiData   & \textbf{0.487} & 0.465\drop{.022} \\
        ARC-Easy             & 0.691 & \textbf{0.716}\gain{.025} \\
        ARC-Challenge        & \textbf{0.403} & 0.396\drop{.007} \\
        \midrule
        \multicolumn{3}{c}{Commonsense reasoning} \\
        \midrule
        COPA                 & \textbf{0.670} & 0.660\drop{.010} \\
        CommonsenseQA        & \textbf{0.334} & 0.324\drop{.010} \\
        PIQA                 & \textbf{0.742} & 0.735\drop{.007} \\
        OpenBookQA           & 0.378 & \textbf{0.380}\gain{.002} \\
        \midrule
        \multicolumn{3}{c}{Language understanding} \\
        \midrule
        LAMBADA              & 0.426 & \textbf{0.451}\gain{.025} \\
        HellaSwag            & 0.558 & \textbf{0.574}\gain{.016} \\
        Winograd             & \textbf{0.659} & 0.645\drop{.014} \\
        Winogrande           & \textbf{0.563} & 0.557\drop{.006} \\
        BIG-bench LangID     & \textbf{0.256} & 0.255\drop{.001} \\
        \bottomrule
    \end{tabular*}
    \caption{Full benchmark comparison of the \textbf{Baseline} and \textbf{BoV} on the 780M model, trained identically on $3.91\times10^{19}$ FLOPs ($\approx$8B tokens).}
    \label{tab:full_results}
\end{table}

%% file: tables/full_results_24l.tex
\begin{table}[t]
    \centering
    \small
    \begin{tabular*}{\columnwidth}{@{\extracolsep{\fill}}lcc@{}}
        \toprule
        & Baseline & BoV \\
        \midrule
        val bpb~$\downarrow$ & 0.671 & \textbf{0.663}\drop{.008} \\
        CORE~$\uparrow$      & 0.334 & \textbf{0.351}\gain{.017} \\
        \midrule
        \multicolumn{3}{c}{Reading comprehension} \\
        \midrule
        SQuAD                & 0.506 & \textbf{0.531}\gain{.025} \\
        CoQA                 & 0.342 & \textbf{0.372}\gain{.030} \\
        BoolQ                & 0.683 & \textbf{0.697}\gain{.014} \\
        \midrule
        \multicolumn{3}{c}{Symbolic problem solving} \\
        \midrule
        BIG-bench Dyck       & 0.087 & \textbf{0.154}\gain{.067} \\
        AGIEval LSAT-AR      & \textbf{0.300} & 0.187\drop{.113} \\
        BIG-bench CS-Alg.\   & \textbf{0.442} & 0.414\drop{.028} \\
        BIG-bench Operators  & 0.210 & \textbf{0.219}\gain{.009} \\
        BIG-bench Rep.-Copy  & 0.000 & \textbf{0.094}\gain{.094} \\
        \midrule
        \multicolumn{3}{c}{World knowledge} \\
        \midrule
        Jeopardy             & 0.219 & \textbf{0.254}\gain{.035} \\
        BIG-bench WikiData   & \textbf{0.568} & 0.561\drop{.007} \\
        ARC-Easy             & 0.743 & \textbf{0.745}\gain{.002} \\
        ARC-Challenge        & \textbf{0.471} & 0.468\drop{.003} \\
        \midrule
        \multicolumn{3}{c}{Commonsense reasoning} \\
        \midrule
        COPA                 & 0.710 & \textbf{0.740}\gain{.030} \\
        CommonsenseQA        & 0.437 & \textbf{0.523}\gain{.086} \\
        PIQA                 & 0.768 & \textbf{0.776}\gain{.008} \\
        OpenBookQA           & 0.416 & \textbf{0.418}\gain{.002} \\
        \midrule
        \multicolumn{3}{c}{Language understanding} \\
        \midrule
        LAMBADA              & 0.504 & \textbf{0.521}\gain{.017} \\
        HellaSwag            & 0.660 & \textbf{0.669}\gain{.009} \\
        Winograd             & \textbf{0.766} & 0.740\drop{.026} \\
        Winogrande           & 0.581 & \textbf{0.609}\gain{.028} \\
        BIG-bench LangID     & 0.246 & \textbf{0.274}\gain{.028} \\
        \bottomrule
    \end{tabular*}
    \caption{Full benchmark comparison of the \textbf{Baseline} and \textbf{BoV} on the 780M model, trained identically on 100B tokens.}
    \label{tab:full_results_24l}
\end{table}

%% file: tables/nanogpt_full.tex
\begin{table}[t]
    \centering
    \small
    \begin{tabular*}{\columnwidth}{@{\extracolsep{\fill}}lcc@{}}
        \toprule
        & $\mathbf{V} + \mathbf{E}_v[i]$ & $\mathbf{E}_v[i]$ \\
        \midrule
        val bpb~$\downarrow$ & \textbf{0.712} & 0.714\gain{.002} \\
        CORE~$\uparrow$      & 0.268 & \textbf{0.272}\gain{.004} \\
        \midrule
        \multicolumn{3}{c}{Reading comprehension} \\
        \midrule
        SQuAD                & \textbf{0.439} & 0.416\drop{.023} \\
        CoQA                 & 0.312 & \textbf{0.321}\gain{.009} \\
        BoolQ                & 0.594 & \textbf{0.635}\gain{.041} \\
        \midrule
        \multicolumn{3}{c}{Symbolic problem solving} \\
        \midrule
        BIG-bench Dyck       & \textbf{0.168} & 0.147\drop{.021} \\
        AGIEval LSAT-AR      & 0.270 & \textbf{0.291}\gain{.021} \\
        BIG-bench CS-Alg.\   & 0.432 & \textbf{0.433}\gain{.001} \\
        BIG-bench Operators  & 0.171 & \textbf{0.195}\gain{.024} \\
        BIG-bench Rep.-Copy  & 0.031 & \textbf{0.063}\gain{.032} \\
        \midrule
        \multicolumn{3}{c}{World knowledge} \\
        \midrule
        Jeopardy             & 0.106 & \textbf{0.119}\gain{.013} \\
        BIG-bench WikiData   & \textbf{0.497} & 0.465\drop{.032} \\
        ARC-Easy             & 0.715 & \textbf{0.716}\gain{.001} \\
        ARC-Challenge        & \textbf{0.400} & 0.396\drop{.004} \\
        \midrule
        \multicolumn{3}{c}{Commonsense reasoning} \\
        \midrule
        COPA                 & 0.660 & 0.660 \\
        CommonsenseQA        & 0.259 & \textbf{0.324}\gain{.065} \\
        PIQA                 & \textbf{0.747} & 0.735\drop{.012} \\
        OpenBookQA           & \textbf{0.386} & 0.380\drop{.006} \\
        \midrule
        \multicolumn{3}{c}{Language understanding} \\
        \midrule
        LAMBADA              & 0.433 & \textbf{0.451}\gain{.018} \\
        HellaSwag            & 0.574 & 0.574 \\
        Winograd             & \textbf{0.692} & 0.645\drop{.047} \\
        Winogrande           & \textbf{0.567} & 0.557\drop{.010} \\
        BIG-bench LangID     & 0.255 & 0.255 \\
        \bottomrule
    \end{tabular*}
    \caption{Full benchmark comparison of the add form ($\mathbf{V} + \mathbf{E}_v[i]$) and the replace form of BoV ($\mathbf{E}_v[i]$) on the 780M model across all 21 DCLM CORE tasks, grouped into the five DCLM categories, plus validation BPB and the aggregate CORE score.}
    \label{tab:nanogpt_full}
\end{table}

%% file: appendices/experimental_details.tex
\section{Additional Experimental Details}
\label{app:details}

Table~\ref{tab:hyperparams} reports the full training configuration for both model scales.

\input{tables/hyperparams}

%% file: tables/hyperparams.tex
\begin{table*}[t]
    \centering
    \small
    \begin{tabular}{lcc}
        \toprule
        & 135M (12-layer) & 780M (24-layer) \\
        \midrule
        \multicolumn{3}{l}{\emph{Architecture}} \\
        Layers & 12 & 24 \\
        Model dimension $d$ & 768 & 1536 \\
        Attention heads ($H$ / KV) & 6 / 6 & 12 / 12 \\
        Head dimension $d_h$ & 128 & 128 \\
        Vocabulary size $|\mathcal{V}|$ & 32{,}768 & 32{,}768 \\
        Window pattern & SSSL & SSSL \\
        Sequence length $T$ & 2048 & 2048 \\
        Parameters & 135.3M & 780.1M \\
        \midrule
        \multicolumn{3}{l}{\emph{Training budget}} \\
        Total batch size (tokens) & 524{,}288 $(2^{19})$ & 1{,}048{,}576 $(2^{20})$ \\
        Device batch size & 16 & 4 \\
        Gradient accumulation (4 GPUs) & 4 & 32 \\
        Iterations & 3{,}766 & 7{,}809 \\
        Total tokens & 1.97B & 8.19B \\
        Target FLOPs & $1.50 \times 10^{18}$ & $3.91 \times 10^{19}$ \\
        FLOPs per token & $7.60 \times 10^{8}$ & $4.78 \times 10^{9}$ \\
        Tokens-to-parameters ratio & 17.9 & 11.2 \\
        Compute dtype & bf16 & bf16 \\
        \midrule
        \multicolumn{3}{l}{\emph{Optimization}} \\
        Optimizer & \multicolumn{2}{c}{Muon (matrices) + AdamW (rest)} \\
        Matrix LR (Muon) & 0.02 & 0.0283 \\
        Embedding LR & 0.30 & 0.30 \\
        Unembedding LR & 0.008 & 0.008 \\
        Scalar LR & 0.50 & 0.707 \\
        Weight decay & 0.28 & 0.0597 \\
        LR schedule & \multicolumn{2}{c}{warmup 40 steps; warmdown from 0.65; final frac.\ 0.05} \\
        \bottomrule
    \end{tabular}
    \caption{Training hyperparameters for the 135M (12-layer) and 780M (24-layer) models. The 780M learning rates and weight decay are scaled from the 135M values following batch-size ($\times\sqrt{2}$) and width scaling rules.}
    \label{tab:hyperparams}
\end{table*}